\documentclass[10pt,conference]{IEEEtran}
\IEEEoverridecommandlockouts
\usepackage{cite}
\usepackage{amsmath,amssymb,amsfonts}
\usepackage{algorithmic}
\usepackage{graphicx}
\usepackage{textcomp}
\usepackage{xcolor}
\usepackage{booktabs}
\def\BibTeX{{\rm B\kern-.05em{\sc i\kern-.025em b}\kern-.08em
    T\kern-.1667em\lower.7ex\hbox{E}\kern-.125emX}}
\begin{document}

\newcommand{\sig}{\textsuperscript{*}}
\newcommand{\nosig}{\phantom{\textsuperscript{*}}}

\title{Budget-Aware Uncertainty for Radiotherapy Segmentation QA Using nnU-Net
}

\author{
\IEEEauthorblockN{Ricardo Coimbra Brioso\IEEEauthorrefmark{1}, 
Lorenzo Mondo\IEEEauthorrefmark{1}, 
Damiano Dei\IEEEauthorrefmark{2}\IEEEauthorrefmark{3}, 
Nicola Lambri\IEEEauthorrefmark{2}\IEEEauthorrefmark{3},\\ 
Pietro Mancosu\IEEEauthorrefmark{3}, 
Marta Scorsetti\IEEEauthorrefmark{2}\IEEEauthorrefmark{3}, 
and Daniele Loiacono\IEEEauthorrefmark{1}}
\smallskip

\IEEEauthorblockA{\IEEEauthorrefmark{1}Dipartimento di Elettronica, Informazione e Bioingegneria, Politecnico di Milano, Milan, Italy}

\IEEEauthorblockA{\IEEEauthorrefmark{2}Department of Biomedical Sciences, Humanitas University, Pieve Emanuele, Milan, Italy}

\IEEEauthorblockA{\IEEEauthorrefmark{3}Radiotherapy and Radiosurgery Department, IRCCS Humanitas Research Hospital, Rozzano, Milan, Italy}

\IEEEauthorblockA{Email: ricardo.brioso@polimi.it, daniele.loiacono@polimi.it}
}

\maketitle

\begin{abstract}
Accurate delineation of the Clinical Target Volume (CTV) is essential for radiotherapy planning, yet remains time-consuming and difficult to assess, especially for complex treatments such as Total Marrow and Lymph Node Irradiation (TMLI). While deep learning--based auto-segmentation can reduce workload, safe clinical deployment requires reliable cues indicating where models may be wrong. In this work, we propose a budget-aware uncertainty-driven quality assurance (QA) framework built on nnU-Net, combining uncertainty quantification and post-hoc calibration to produce voxel-wise uncertainty maps (based on predictive entropy) that can guide targeted manual review. We compare temperature scaling (TS), deep ensembles (DE), checkpoint ensembles (CE), and test-time augmentation (TTA), evaluated both individually and in combination on TMLI as a representative use case. Reliability is assessed through ROI-masked calibration metrics and uncertainty--error alignment under realistic revision constraints, summarized as AUC over the top 0--5\% most uncertain voxels. Across configurations, segmentation accuracy remains stable, whereas TS substantially improves calibration. Uncertainty--error alignment improves most with calibrated checkpoint-based inference, leading to uncertainty maps that highlight more consistently regions requiring manual edits. Overall, integrating calibration with efficient ensembling seems a promising strategy to implement a budget-aware QA workflow for radiotherapy segmentation. 
\end{abstract}


\section{Introduction}
\label{sec:introduction}
Radiotherapy is used in more than 50\% of cancer patients, as a primary treatment or within multimodal care~\cite{Baskar2012CancerAR}. Treatment planning requires delineating the Clinical Target Volume (CTV) and organs at risk (OARs) on CT to deliver ionizing radiation dose to the target while sparing healthy tissues, a process that remains time-consuming and error-prone. Over the last decade, deep learning (DL) has enabled accurate automatic contouring for many common structures~\cite{huynh2020artificial,cardenas2019advances,shi2022deep}; however, clinical adoption is still limited for complex \emph{targets}, where reliability and interpretability are as critical as average segmentation accuracy.

This limitation is particularly evident in Total Marrow and Lymph Node Irradiation (TMLI)~\cite{mancosu2020}, where targets are large, spatially extended, and heterogeneous (e.g., marrow compartments and lymph-node chains). In this setting, clinically relevant errors can be subtle yet consequential: omissions may lead to underdosage and risk of recurrence, while over-contouring can increase widespread toxicity due to the large treated volume. Although automatic segmentation can reduce workload, full-body CT scans span hundreds of slices and multiple anatomical regions, making uniform and exhaustive manual verification unrealistic within routine clinical constraints. Therefore, the key practical question becomes how to prioritize review towards regions where the model is most likely wrong.

Compared to OARs, complex targets such as CTVs often present weaker and ambiguous boundaries on CT, higher inter-patient variability, and greater inter-observer disagreement; as a result, a model may output seemingly confident contours even when clinically relevant regions are missed~\cite{Wahid2024,Dnger2025}. Uncertainty quantification (UQ) can mitigate this trust gap by associating each voxel prediction with a confidence estimate and visualizing it as an uncertainty map. Yet, many UQ works in medical segmentation emphasize global calibration and scoring rules (e.g., NLL/Brier/ECE) without assessing whether uncertainty is \emph{spatially actionable} for quality assurance (QA), i.e., whether it highlights true errors under realistic, budget-limited review. In addition, computationally expensive choices (e.g., deep ensembles) are often used as references without a systematic comparison against practical alternatives (e.g., checkpoint ensembles, test-time augmentation) within the same QA objective.

In this work, we aim at investigating whether combining complementary UQ strategies with post-hoc calibration yields uncertainty maps that better align with segmentation errors and therefore better support clinical QA. Building on a strong nnU-Net baseline for TMLI target segmentation~\cite{isensee2021nnunet,brioso2024investigatinggenderbiaslymphnode}, we compare a single segmentation model, deep ensembles, and checkpoint ensembles, each optionally combined with test-time augmentation and temperature scaling, and we produce predictive-entropy maps to guide review. To evaluate clinical utility, we introduce ROI-masked, budget-based uncertainty--error alignment analyses over constrained uncertain-volume budgets, and we illustrate how threshold/budget operating points can generate uncertainty overlays that qualitatively match clinician corrections and enable progressive, time-aware inspection.

% Section~\ref{sec:methods} describes data, model, and UQ/calibration configurations; Section~\ref{sec:results} reports accuracy, calibration, and budget-aware alignment; Section~\ref{sec:discussion} discusses implications and limitations, followed by conclusions in Section~\ref{sec:conclusions}.

\section{Related Work}
\label{sec:related}
\subsection{Deep learning for complex target delineation in radiotherapy}
DL is increasingly adopted for auto-contouring in RT, with benefits in efficiency and contour consistency, especially for organs-at-risk (OARs)~\cite{wong2021implementation,harrison2022machine}.
Isensee et al.~\cite{isensee2021nnunet} introduced the nnU-Net framework, which has enabled robust 3D segmentation pipelines and large-scale RT auto-contouring systems~\cite{Tang2019,CHEN2021175,Wasserthal2022}.
Despite the proliferation of architectural variants, Isensee et al. showed that nnU-Net remains highly competitive when assessed under rigorous validation protocols, supporting its broad adoption as a reliable baseline~\cite{Isensee2024}.

However, clinical target volumes (CTV), especially complex ones like lymphatic targets (CTV$_\mathrm{LN}$), remain more challenging than OARs.
Sharp et al.~\cite{sharp2014vision} discussed how guideline-driven definitions and limited image contrast increase inter-/intra-observer variability and create local failure modes that require careful manual review, while Kalantar et al.~\cite{Kalantar2021} highlighted the impact of anatomical variability and ambiguous boundaries on pelvic target delineation.
In TMLI, Wong et al. introduced targeted marrow/lymphoid irradiation as an alternative to conventional TBI, and Mancosu et al. reviewed its technological and clinical development, emphasizing the practical challenges of large-field target definition~\cite{Wong2006,mancosu2020}.
Dei et al.~\cite{Dei2023} showed that internal contouring guidelines can reduce variability for lymph node targets in TMLI, underscoring both the complexity of CTV$_\mathrm{LN}$ and the value of protocol-driven standardization.

% Several DL strategies have been explored for lymphatic target delineation.
% Shi et al.~\cite{shi2022} proposed a Transformer--CNN dual-encoder architecture for automatic segmentation of TMLI target structures.
% Li and Xia~\cite{Li2021} used weak supervision and reinforcement learning to refine coarse annotations into voxel-wise lymph node segmentations.
% Bouget et al.~\cite{Bouget2023} incorporated anatomical priors and ensemble-style designs to improve robustness in lymph node segmentation.
% Costea et al.~\cite{Costea2023} clinically evaluated head-and-neck lymph node segmentation algorithms, and Mikalsen et al.~\cite{Mikalsen2023} reported extensive clinical testing of RT segmentation models, both indicating that complex targets more often require manual correction than standard OARs.

\subsection{Uncertainty quantification and visualization for clinical review}
Reliable clinical deployment requires signals of \emph{when and where} a model may fail.
Guo et al.~\cite{guo2017calibration} demonstrated that modern neural networks are often miscalibrated and proposed temperature scaling as a simple post-hoc calibration method.
Mehrtash et al.~\cite{mehrtash2020confidence} and Jungo et al.~\cite{Jungo2020} analyzed calibration and predictive uncertainty in medical image segmentation, noting that voxel-wise uncertainty can be noisy and that subject-level reliability may degrade even when dataset-level calibration appears acceptable.

UQ methods include epistemic approximations via deep ensembles~\cite{lakshminarayanan2017simple} and efficient posterior sampling such as checkpoint ensembles~\cite{zhao2022efficient}.
Wang et al.~\cite{wang2019aleatoric} used test-time augmentation (TTA) to probe input-related ambiguity via prediction variability, while Baumgartner et al.~\cite{baumgartner2019phiseg} introduced PHiSeg to model multiple plausible segmentations in ambiguous settings.
At the evaluation level, Wahid et al.~\cite{Wahid2024} reviewed UQ in RT and noted that head-to-head comparisons and calibration are still relatively uncommon, and Kahl et al.~\cite{Kahl2024} proposed ValUES to systematize validation across downstream tasks and emphasize the role of ablations and score aggregation.

Clinical usefulness depends on intuitive visualization and actionable summaries.
De Biase et al.~\cite{DeBiase2024} designed and assessed a GUI for tumor probability maps, reporting that radiation oncologists (ROs) value threshold sliders and controls to suppress low-importance values without discarding information.
Rodr\'iguez Outeiral et al.~\cite{RodrguezOuteiral2023} proposed score-based QA metrics for review triage, and De Biase et al.~\cite{DeBiase2025} showed that uncertainty/probability-map statistics can correlate with accuracy in external cohorts.
Maruccio et al.~\cite{Maruccio2024} and Rogowski et al.~\cite{Rogowski2025} reported that uncertainty information can support correction workflows by highlighting critical regions and reducing editing time when presented in an interpretable manner.
Finally, van Aalst et al.~\cite{vanAalst2025} benchmarked UQ for head-and-neck OAR auto-segmentation with a 3D U-Net within nnU-Net, comparing MC dropout, deep ensembles, and TTA, and motivating careful method comparison, calibration, and validation against RO edits for complex targets such as CTV$_\mathrm{LN}$.

\section{Uncertainty Estimation for nnU-Net}
\label{sec:methods}
To address the gaps identified in the literature, we integrated multiple uncertainty-estimation strategies into the nnU-Net framework. Following the ValUES framework \cite{Kahl2024}, we combine complementary approaches in order to (i) calibrate predicted probabilities and (ii) account for prediction variability.
All methods considered in this work ultimately yield a voxel-wise probability map. 
We quantify voxel-level predictive uncertainty using entropy computed from these probabilities, as typically done in medical image segmentation \cite{zhao2022efficient,wang2019aleatoric,Kahl2024,Maruccio2024}.
For the case of binary segmentation, considered in this work, we let $p_i \in [0,1]$ denote the predicted probability of the foreground class at voxel $i$ and compute uncertainty as the entropy of the corresponding Bernoulli distribution as follows.
\begin{equation}
    H(p_i) = -p_i \log_2(p_i + \epsilon) - (1 - p_i)\log_2(1 - p_i + \epsilon),
\label{eq:entropy_visualization}
\end{equation}
where $\epsilon$ is a small constant (e.g., $10^{-10}$) added for numerical stability. 
When employing methods producing multiple probabilistic predictions per voxel (e.g., ensembles or TTA), we first aggregate them into a mean probability map $\overline{p}_i$ and then compute $H(\overline{p}_i)$.

\subsection{Temperature scaling}
We apply temperature scaling to modulate the sharpness of the softmax distribution and, consequently, the confidence of the predicted probabilities. 
Given logits $z_{i,c}$ for voxel $i$ and class $c$, temperature scaling adjusts the predicted probabilities as follows:
\begin{equation}
    p_{i,c} = \frac{\exp(z_{i,c}/T)}{\sum_{j}\exp(z_{i,j}/T)}.
\label{eq:temperature_scaling}
\end{equation}
Larger values of $T$ yield softer (less peaked) probability distributions, while $T<1$ produces sharper outputs. 
In our implementation, $T$ is exposed as a configurable inference-time parameter within the nnU-Net pipeline.

\subsection{Ensemble methods}
Ensembling captures predictive variability by aggregating multiple probabilistic segmentations. Given $M$ models (or model instances) producing voxel-wise probabilities $p_{i}^{(m)}$, we compute the mean probability
\[
\overline{p}_i = \frac{1}{M}\sum_{m=1}^{M} p_{i}^{(m)},
\]
and derive voxel-wise uncertainty via $H(\overline{p}_i)$.
In this work, we explored two different ensembling strategies: deep ensembles and checkpoint ensembles.

\subsubsection{Deep ensembles}
We use the nnU-Net cross-validation models as a deep ensemble. During inference, the final voxel-wise probability map is obtained by averaging the softmax outputs across the $M$ independently trained folds, yielding $\overline{p}_i$.

\subsubsection{Checkpoint ensembles}
Checkpoint ensembling estimates predictive variability by aggregating model states saved at different points of a single training run. 
We adopt the cyclical strategy described in \cite{Zhao2024}, training for a total of $E_{\mathrm{tot}}=1200$ epochs using $3$ learning-rate cycles of length $T_c=400$ epochs. 
The learning rate $\alpha(t)$ at epoch $t$ within a cycle ($t_c$) follows
\begin{equation}
    \alpha(t) =
\begin{cases}
  \alpha_{r} & t_c=0 \\
  \alpha_{0}\left[1-\frac{\min(t_c, \gamma T_c)}{T_c}\right]^{\epsilon} & t_c\neq0,
\end{cases}
\label{eq:cyclical_learningrate}
\end{equation}
with $\alpha_r=0.1$, $\alpha_0=0.01$, $\gamma=0.8$, and $\epsilon=0.9$.
During the final 10 epochs of each cycle, we saved one checkpoint per epoch (30 checkpoints in total). 
At inference time, we aggregate predictions from a subset of these checkpoints (5 in this work) by averaging their probability maps to obtain $\overline{p}_i$, from which entropy-based uncertainty is computed.

\subsection{Test-time augmentation}
We use TTTA by applying $N$ random, small-magnitude transformations to the input volume, performing inference on each transformed input, and mapping predictions back to the original space. We used rotations in $[-5^\circ, +5^\circ]$, translations in $[-5, +5]$ voxels, and intensity scaling in $[0.9, 1.1]$. 
The final probability map is computed as the mean of the de-augmented predictions, yielding $\overline{p}_i$, and voxel-wise uncertainty is obtained as $H(\overline{p}_i)$.

\section{Experimental Design}
\label{sec:design}
\subsection{Dataset and Segmentation Model}
\label{ssec:dataset}
In this paper, we focused on the segmentation of lymph nodes and the spleen in CT images acquired for TMLI treatment planning. In particular, the patients included in our dataset were diagnosed with hematological malignancies, identified as candidates for allogeneic transplantation, and treated with non-myeloablative TMLI at \emph{AnonymousHospital}.
%Humanitas Research Hospital (Rozzano, Milan, Italy).
All TMLI patients were immobilized in the supine position with their arms alongside the body using an in-house dedicated frame with multiple personalized masks~\cite{MANCOSU2021e98}. A free-breathing, non-contrast CT scan was acquired for each patient using a BigBore CT system (Philips Healthcare, Best, The Netherlands), with a slice thickness of 5~mm.
The average CT volume shape is $237 \times 512 \times 512$, with an average voxel spacing of $5.0~\mathrm{mm} \times 1.171875~\mathrm{mm} \times 1.171875~\mathrm{mm}$. Following nnU-Net preprocessing, intensity statistics (mean, standard deviation, and the $0.5$/$99.5$ percentiles) are computed on the global target class. Intensities are clipped to the percentile range and then standardized by subtracting the mean and dividing by the standard deviation.

In TMLI, the target structure is defined as the union of bone marrow (CTV\_BM), spleen (CTV\_Spleen), and all lymph node chains (CTV\_LN). The CTV\_BM was considered equivalent to the skeletal bones, with the chest wall added to the ribs to account for breathing motion. To minimize oral cavity toxicity, the mandible was excluded from CTV\_BM, along with the hands, which have an extremely limited bone marrow presence.
The total planning target volume (PTV\_Tot) was defined as the union of three PTVs derived from isotropic expansions of the corresponding CTVs: (i) PTV\_BM = CTV\_BM + 2~mm (+8~mm for arms and legs) to account for setup margin; (ii) PTV\_Spleen = CTV\_Spleen + 5~mm to account for breathing motion and setup margin; and (iii) PTV\_LN = CTV\_LN + 5~mm to account for target residual motion and setup margin.
In this work, we defined the segmentation target as the union of CTV\_LN and CTV\_Spleen, which are among the most challenging structures to segment automatically due to their high inter-patient variability and complex shapes.

Overall, a cohort of 45 patients, consisting of 25 male and 19 female patients, was available.
Our experimental setup was designed to build a robust training and evaluation set from the full cohort of 45 patients.
We employed a repeated hold-out strategy to create four distinct train/test splits.
For each split, 5 patients were held out to assess model performance and uncertainty estimates, and the remaining 40 patients were used to train the segmentation models.
A standard 5-fold cross-validation was employed, yielding 5 training folds of 32 patients and 5 validation folds of 8 patients each.
These five folds were used to train the five models required for the Deep Ensemble strategy (see Section~\ref{sec:methods}); in contrast, when a single model or checkpoint ensembling was employed, only one training/validation fold pair was used.
% Finally, an additional set of 8 patients was used for prospective clinical evaluation, as described in more detail in Section~\ref{ssec:additional_results}.

\subsection{Metrics}
\label{sec:uncertainty_metrics}
In this section, we present the quantitative and qualitative methods used to assess segmentation quality, calibration, and the ability of voxel-wise uncertainty to localize prediction errors.

\paragraph{Calibration metrics (ROI-masked)}
Voxel-wise calibration metrics can be dominated by the large number of True Negative background voxels. To focus the evaluation on clinically relevant areas, we compute calibration metrics within a region of interest (ROI), denoted as $\mathcal{R}$. Specifically, $\mathcal{R}$ is defined as the union of voxels within a fixed distance $\delta$ (set to 15~mm) from either the predicted boundary $\partial A$ or the ground truth boundary $\partial B$:
\[
\mathcal{R} = \{ i \mid d(i, \partial A) \le \delta \} \cup \{ i \mid d(i, \partial B) \le \delta \}.
\]
Unless otherwise stated, all calibration results in this work refer to metrics computed over voxels $i \in \mathcal{R}$.

We report the Expected Calibration Error (ECE). Predictions are divided into $M$ bins $\{B_m\}_{m=1}^M$ according to their confidence $\hat{p}_i$, and we restrict each bin to the ROI, $B_m^{\mathcal{R}} = B_m \cap \mathcal{R}$. The accuracy and confidence of each (restricted) bin are defined as
\begin{equation}
\begin{split}
\mathrm{acc}(B_m^{\mathcal{R}}) &= \frac{1}{|B_m^{\mathcal{R}}|}
  \sum_{i \in B_m^{\mathcal{R}}} \mathbf{1}(\hat{y}_i = y_i),\\
\mathrm{conf}(B_m^{\mathcal{R}}) &= \frac{1}{|B_m^{\mathcal{R}}|}
  \sum_{i \in B_m^{\mathcal{R}}} \hat{p}_i,
\end{split}
\label{eq:accuracy_confidence}
\end{equation}

and the ECE is computed as
\begin{equation}
    \mathrm{ECE} = \sum_{m=1}^{M} \frac{|B_{m}^{\mathcal{R}}|}{|\mathcal{R}|}
    \left| \mathrm{acc}(B_{m}^{\mathcal{R}}) - \mathrm{conf}(B_{m}^{\mathcal{R}}) \right|.
    \label{eq:ECE}
\end{equation}

The Brier Score (BS) quantifies the accuracy of probabilistic predictions by measuring the mean squared error between predicted probabilities and ground truth labels, computed over the same ROI:
\begin{equation}
    \mathrm{BS} = \frac{1}{|\mathcal{R}|} \sum_{i \in \mathcal{R}} (\hat{p}_i - y_i)^2.
    \label{eq:brier_score}
\end{equation}

\paragraph{Segmentation accuracy}
The Dice Similarity Coefficient (DSC) compares the similarity between the predicted binary segmentation and the ground truth:
\begin{equation}
    \mathrm{DSC} = \frac{2 \times |A \cap B|}{|A| + |B|},
    \label{eq:dice_score}
\end{equation}
where $A$ is the predicted set of voxels and $B$ is the ground truth set.

\paragraph{Uncertainty--error overlap}
The Uncertainty--Error Overlap (UEO) assesses how well the spatial distribution of model uncertainty aligns with segmentation errors, as introduced in~\cite{Jungo2020}. Let $u_i$ denote a voxel-wise uncertainty score (entropy, computed from Eq.~\ref{eq:entropy_visualization}). For a threshold $\tau$, define the uncertain set $U_{\tau} = \{ i \mid u_i > \tau \}$ and the error set $E = \mathrm{FP} \cup \mathrm{FN}$. The UEO is the Dice overlap between $U_{\tau}$ and $E$:
\begin{equation}
    \mathrm{UEO}(\tau) = \frac{2 \times |U_{\tau} \cap E|}{|U_{\tau}| + |E|}.
    \label{eq:ueo_dice}
\end{equation}

\paragraph{Percentile-based AUC metrics under a revision budget.}
To quantify how effectively voxel-wise uncertainty highlights segmentation errors under a limited revision budget, we threshold uncertainty maps by selecting a fixed fraction of voxels as ``uncertain'' and compute overlap-based metrics on the resulting binary masks. Importantly, the operating point is defined via percentiles of the uncertainty distribution (rather than a fixed uncertainty threshold), which enables fair comparisons across methods with different uncertainty scaling.

For each volume and each budget $b \in [v_1, v_2]\%$, we define the set of flagged voxels $U_b$ as the top-$b\%$ most uncertain voxels. Equivalently, if $\tau_b$ denotes the $(100-b)$-th percentile of the uncertainty values $\{u_i\}$ in that volume, then
\[
U_b = \{ i \mid u_i \ge \tau_b \}.
\]
We then compute a metric value $\mathrm{metric}(b)$ on $U_b$ (e.g., UEO computed by replacing $U_{\tau}$ with $U_b$ in Eq.~\ref{eq:ueo_dice}).

When the percentile threshold falls within a plateau (multiple voxels sharing the same uncertainty value at the boundary), the selection of exactly $b\%$ voxels is not deterministic. In this case, we randomly sample the required number of voxels from the plateau, repeat the procedure $S=10$ times, and average the resulting metric values. When no plateau is present, the selection is deterministic and the metric is computed once.

Finally, for each case we obtain a curve $\mathrm{metric}(b)$ over $b \in [v_1, v_2]\%$ and summarize it by the area under the curve (AUC), computed by trapezoidal integration and normalized by the interval length:
\[
\mathrm{metric}_{\mathrm{AUC}} = \frac{1}{v_2 - v_1}\int_{v_1}^{v_2} \mathrm{metric}(b)\, db.
\]

We report three AUC-based metrics:
\begin{enumerate}
    \item \textbf{$\mathrm{UEO}_{\mathrm{AUC}}$:} AUC of $\mathrm{UEO}(b)$ computed using $U_b$.
    \item \textbf{$(\mathrm{FP}-\mathrm{TP})_{\mathrm{AUC}}$:} AUC of $\mathrm{Cov}_{\mathrm{FP}}(b) - \mathrm{Cov}_{\mathrm{TP}}(b)$, where coverage is the fraction of voxels of a given set flagged as uncertain:
    \[
    \mathrm{Cov}_{C}(b) = \frac{|U_b \cap C|}{|C|}, \qquad C \in \{\mathrm{FP}, \mathrm{TP}\}.
    \]
    Higher values indicate that false positives are preferentially flagged as uncertain while true positives remain comparatively confident.
    \item \textbf{$(\mathrm{FN}-\mathrm{TN})_{\mathrm{AUC}}$:} AUC of $\mathrm{Cov}_{\mathrm{FN}}(b) - \mathrm{Cov}_{\mathrm{TN}}(b)$, with $\mathrm{Cov}_{C}(b)$ defined analogously for $C \in \{\mathrm{FN}, \mathrm{TN}\}$. Higher values indicate that false negatives are preferentially flagged while true negatives remain comparatively confident.
\end{enumerate}

\subsection{Statistical analysis}
To assess whether observed differences between experiments were statistically supported, we performed a nonparametric analysis on the patient-by-experiment matrix built for each metric. We applied the Friedman test to detect an overall effect of the experimental condition. After the global test, we carried out pairwise comparisons between all pairs of experiments using the Wilcoxon signed-rank test, and corrected the resulting $p$-values with the Benjamini--Hochberg false discovery rate (FDR) procedure to control for multiple testing. Finally, for each metric we compared every experiment against the best-performing one and marked with a superscript ``$^*$'' those entries whose FDR-adjusted $p$-value was below $\alpha = 0.05$.

\section{Results and Discussion}
\label{sec:results}
We evaluated segmentation accuracy, calibration, and uncertainty--error alignment across different uncertainty estimation strategies, considered both individually and in combination. 
We employed a repeated hold-out train/test procedure with four repetitions (see Section~\ref{ssec:dataset} for more details). 
For each repetition, we trained: (i) a single baseline model (\texttt{BASE}); (ii) a Deep Ensemble (\texttt{DE}) composed of five models trained via 5-fold cross-validation; and (iii) a Checkpoint Ensemble (\texttt{CE}) composed of five checkpoints saved during the training of a single model (see Section~\ref{sec:methods} for more details). 
Each of these three approaches was evaluated both alone and in combination with \texttt{TTA} only, \texttt{TS} only, or with both \texttt{TTA} and \texttt{TS}, resulting in 12 configurations in total. \texttt{TTA} was implemented using five augmentations, and \texttt{TS} used a fixed temperature $T=3$; both hyperparameters were selected based on preliminary analyses.

\subsection{Experimental results}
\label{ssec:results_main}
Table~\ref{tab:uncertainty_metrics_summary}, summarize the main segmentation and calibration results across all evaluated configurations. Each configuration was evaluated using 4 different train/test splits (repeated hold-out), and the reported metrics represent the mean and standard deviation across the 20 patients of the 4 test splits. 

The results suggest that ensembling improves DSC over the single-model baseline. The best mean DSC is achieved by \texttt{DE} (and its variants), while \texttt{CE} remains close; however, post-hoc results show that several ensemble-based configurations are not significantly different from the best, suggesting that DSC differences are small and often not robust on this dataset. Adding \texttt{TTA} on top of ensembles yields marginal changes and, for some \texttt{CE} configurations (e.g., \texttt{CE+TS+TTA}), DSC becomes significantly lower than the best method, suggesting diminishing returns (or mild degradation) when stacking multiple techniques.

In contrast, calibration benefits more consistently from \texttt{TS}. Across configurations, temperature scaling reduces ECE substantially and several \texttt{TS} variants are not significantly different from the best ECE, confirming that calibration can be improved with negligible additional training cost and without affecting the hard segmentation (as expected). For the Brier score, the advantage concentrates on \texttt{DE+TS} (best BS and not significantly different from the best), while many non-\texttt{TS} variants remain significantly worse, suggesting that combining \texttt{TS} with ensembling yields the most consistent probabilistic improvements.

\begin{table}
\caption{Summary of DSC and ROI-masked calibration metrics (mean $\pm$ std); \textbf{bold} indicates the best method per metric; a global Friedman test across the 12 methods showed significant differences for all metrics ($p<0.0001$); \sig denotes results significantly different from the best method after FDR correction ($\alpha=0.05$).}
\centering
\resizebox{\columnwidth}{!}{
\begin{tabular}{lccc}
\toprule
Method & DSC $\uparrow$ & ECE $\downarrow$ & BS $\downarrow$ \\
\midrule
BASE & 0.838 $\pm$ 0.046\sig & 0.062 $\pm$ 0.015\sig & 0.068 $\pm$ 0.014\sig \\
BASE+TS & 0.838 $\pm$ 0.046\sig & 0.031 $\pm$ 0.006\nosig & 0.059 $\pm$ 0.011\sig \\
BASE+TTA & 0.837 $\pm$ 0.045\sig & 0.042 $\pm$ 0.015\sig & 0.063 $\pm$ 0.013\sig \\
BASE+TS+TTA & 0.837 $\pm$ 0.045\sig & 0.034 $\pm$ 0.009\nosig & 0.059 $\pm$ 0.010\sig \\
DE & \textbf{0.843 $\pm$ 0.044}\nosig & 0.058 $\pm$ 0.015\sig & 0.065 $\pm$ 0.014\sig \\
DE+TS & 0.843 $\pm$ 0.044\nosig & \textbf{0.031 $\pm$ 0.008}\nosig & \textbf{0.056 $\pm$ 0.011}\nosig \\
DE+TTA & 0.842 $\pm$ 0.043\nosig & 0.041 $\pm$ 0.015\sig & 0.061 $\pm$ 0.013\sig \\
DE+TS+TTA & 0.842 $\pm$ 0.043\nosig & 0.034 $\pm$ 0.011\nosig & 0.057 $\pm$ 0.010\nosig \\
CE & 0.840 $\pm$ 0.043\nosig & 0.048 $\pm$ 0.015\sig & 0.063 $\pm$ 0.013\sig \\
CE+TS & 0.840 $\pm$ 0.043\nosig & 0.044 $\pm$ 0.011\sig & 0.059 $\pm$ 0.010\sig \\
CE+TTA & 0.837 $\pm$ 0.041\sig & 0.041 $\pm$ 0.014\sig & 0.062 $\pm$ 0.012\sig \\
CE+TS+TTA & 0.836 $\pm$ 0.041\sig & 0.048 $\pm$ 0.013\sig & 0.060 $\pm$ 0.009\sig \\
\bottomrule
\end{tabular}
}
\label{tab:uncertainty_metrics_summary}
\end{table}

Table~\ref{tab:auc_metrics_summary} reports AUC-based uncertainty--error alignment metrics over the $[0,5]\%$ uncertain-volume range. The choice of focusing on the top 5\% most uncertain voxels reflects a clinically motivated scenario in which only a limited portion of the image can realistically be forwarded to manual quality assurance; reviewing larger fractions rapidly becomes time-consuming with diminishing returns. In our use case (TMLI CTV segmentation for radiotherapy planning), 5\% represents a reasonable upper bound, given that the targets occupy a small fraction of the total volume. Our results show a partial decoupling between segmentation accuracy and uncertainty quality: configurations including \texttt{CE} and/or \texttt{TTA} tend to provide higher $UEO_{\mathrm{AUC}}$, with \texttt{CE+TS+TTA} achieving the best value. Similarly, \texttt{CE+TTA} maximizes $(\mathrm{FP}-\mathrm{TP})_{\mathrm{AUC}}$, while \texttt{DE+TS+TTA} maximizes $(\mathrm{FN}-\mathrm{TN})_{\mathrm{AUC}}$. These results suggest that (i) checkpoint-based diversity and test-time stochasticity can strengthen the spatial concentration of uncertainty on errors, and (ii) the optimal configuration may depend on whether the clinical priority is preferentially flagging false positives or false negatives.

\begin{table}
\caption{Summary of AUC metrics (mean $\pm$ std); \textbf{bold} indicates the best method per metric; a global Friedman test across the 12 methods showed significant differences for all metrics ($p<0.0001$); \sig denotes results significantly different from the best method after FDR correction ($\alpha=0.05$).}
\centering
\resizebox{\columnwidth}{!}{%
\begin{tabular}{lccc}
\toprule
Method & $UEO_{AUC}$ & $(FP-TP)_{AUC}$ & $(FN-TN)_{AUC}$ \\
\midrule
BASE & 0.154 $\pm$ 0.037\sig & 0.057 $\pm$ 0.023\sig & 0.674 $\pm$ 0.092\sig \\
BASE+TS & 0.170 $\pm$ 0.040\nosig & 0.067 $\pm$ 0.027\sig & 0.876 $\pm$ 0.043\sig \\
BASE+TTA & 0.169 $\pm$ 0.039\sig & 0.078 $\pm$ 0.031\sig & 0.886 $\pm$ 0.040\sig \\
BASE+TS+TTA & 0.170 $\pm$ 0.039\nosig & 0.072 $\pm$ 0.029\sig & 0.888 $\pm$ 0.038\sig \\
DE & 0.152 $\pm$ 0.038\sig & 0.059 $\pm$ 0.026\sig & 0.701 $\pm$ 0.081\sig \\
DE+TS & 0.167 $\pm$ 0.042\sig & 0.078 $\pm$ 0.032\sig & 0.894 $\pm$ 0.029\sig \\
DE+TTA & 0.166 $\pm$ 0.040\sig & 0.087 $\pm$ 0.033\sig & 0.897 $\pm$ 0.029\sig \\
DE+TS+TTA & 0.167 $\pm$ 0.040\sig & 0.083 $\pm$ 0.033\sig & \textbf{0.900 $\pm$ 0.027}\nosig \\
CE & 0.171 $\pm$ 0.041\nosig & 0.096 $\pm$ 0.034\sig & 0.898 $\pm$ 0.031\nosig \\
CE+TS & 0.172 $\pm$ 0.040\nosig & 0.092 $\pm$ 0.033\sig & 0.898 $\pm$ 0.032\nosig \\
CE+TTA & 0.171 $\pm$ 0.038\sig & \textbf{0.101 $\pm$ 0.031}\nosig & 0.896 $\pm$ 0.033\nosig \\
CE+TS+TTA & \textbf{0.173 $\pm$ 0.039}\nosig & 0.099 $\pm$ 0.032\sig & 0.899 $\pm$ 0.032\nosig \\
\bottomrule
\end{tabular}%
\label{tab:auc_metrics_summary}
}
\end{table}

Point-wise results at fixed uncertain-volume budgets (Table~\ref{tab:uncertain-volume-metrics}) further clarify the practical trade-offs, especially at tight budgets. At 0.5\% uncertain volume, \texttt{CE+TS+TTA} achieves the highest $UEO$, indicating the strongest local uncertainty--error alignment under severe review constraints, while \texttt{DE+TS} yields the best $FP_{\mathrm{coverage}}$ and \texttt{DE+TS+TTA} the best $FN_{\mathrm{coverage}}$. As the budget increases (2--5\%), coverage metrics approach saturation (e.g., $FP_{\mathrm{coverage}} \approx 1$ for most methods), suggesting diminishing returns and reduced discriminability among configurations at larger review fractions.

Overall, these findings support a pragmatic view: \texttt{TS} is a cost-effective component to improve calibration across all backbones; \texttt{DE} offers the most consistent (though often modest) gains in DSC; and \texttt{CE}/\texttt{TTA} combinations tend to yield more informative uncertainty maps for uncertainty-guided quality assurance, particularly under small uncertain-volume budgets.

\begin{table*}
\caption{Point-wise summary of $UEO$, $FP_{\mathrm{coverage}}$, and $FN_{\mathrm{coverage}}$ at 0.5\%, 1\%, 2\%, and 5\% uncertain-volume budgets; \textbf{bold} indicates the best value per metric--percentage column; $^*$ denotes results significantly different from the best one after FDR correction ($\alpha=0.05$).}    
\centering
\begin{tabular}{lllll|llll|llll}
\toprule
\multicolumn{1}{c}{} & \multicolumn{4}{c|}{$UEO$} & \multicolumn{4}{c|}{$FP_{coverage}$} & \multicolumn{4}{c}{$FN_{coverage}$}\\
\multicolumn{1}{c}{Method} 
& \multicolumn{1}{c}{0.5\%} & \multicolumn{1}{c}{1\%} & \multicolumn{1}{c}{2\%} & \multicolumn{1}{c|}{5\%}
& \multicolumn{1}{c}{0.5\%} & \multicolumn{1}{c}{1\%} & \multicolumn{1}{c}{2\%} & \multicolumn{1}{c|}{5\%}
& \multicolumn{1}{c}{0.5\%} & \multicolumn{1}{c}{1\%} & \multicolumn{1}{c}{2\%} & \multicolumn{1}{c}{5\%} \\ 
\midrule

BASE 
& 0.374\sig & 0.240\sig & 0.131\sig & 0.056\nosig 
& 0.887\sig & 0.991\nosig & 1.000\nosig & 1.000\nosig
& 0.687\sig & 0.718\sig & 0.721\sig & 0.729\sig \\

BASE+TS 
& 0.383\nosig & 0.266\nosig & 0.154\nosig & 0.066\nosig
& 0.897\sig & 0.995\nosig & 1.000\nosig & 1.000\nosig
& 0.716\sig & 0.881\sig & 0.969\sig & 0.978\sig \\

BASE+TTA 
& 0.379\sig & 0.268\nosig & 0.155\nosig & 0.066\nosig
& 0.867\sig & 0.992\sig & 1.000\nosig & 1.000\nosig
& 0.731\sig & 0.896\sig & 0.977\sig & 0.986\sig \\

BASE+TS+TTA 
& 0.386\nosig & 0.267\nosig & 0.155\nosig & 0.066\nosig
& 0.890\sig & 0.995\nosig & 1.000\nosig & 1.000\nosig
& 0.739\sig & 0.895\sig & 0.979\sig & 0.988\sig \\

DE 
& 0.374\sig & 0.236\sig & 0.129\sig & 0.055\nosig
& 0.906\nosig & 0.994\nosig & 1.000\nosig & 1.000\nosig
& 0.715\sig & 0.747\sig & 0.749\sig & 0.757\sig \\

DE+TS 
& 0.384\nosig & 0.262\sig & 0.150\sig & 0.064\nosig
& \textbf{0.908}\nosig & \textbf{0.996}\nosig & 1.000\nosig & 1.000\nosig
& 0.752\nosig & 0.910\nosig & 0.982\sig & 0.990\sig \\

DE+TTA 
& 0.374\sig & 0.263\sig & 0.151\sig & 0.064\nosig
& 0.879\sig & 0.993\sig & 1.000\nosig & 1.000\nosig
& 0.745\sig & 0.919\nosig & 0.988\nosig & 0.993\nosig \\

DE+TS+TTA 
& 0.383\sig & 0.263\sig & 0.151\sig & 0.064\nosig
& 0.901\sig & 0.996\nosig & 1.000\nosig & 1.000\nosig
& \textbf{0.760}\nosig & 0.918\nosig & \textbf{0.989}\nosig & \textbf{0.995}\nosig \\

CE 
& 0.387\nosig & 0.271\nosig & 0.155\nosig & 0.066\nosig
& 0.871\sig & 0.989\sig & 1.000\nosig & 1.000\nosig
& 0.754\nosig & 0.917\nosig & 0.984\nosig & 0.995\nosig \\

CE+TS 
& 0.392\nosig & 0.271\nosig & 0.155\nosig & 0.066\nosig
& 0.886\sig & 0.991\sig & 1.000\nosig & 1.000\nosig
& 0.758\nosig & 0.917\sig & 0.984\nosig & 0.994\nosig \\

CE+TTA 
& 0.384\sig & 0.273\nosig & 0.156\nosig & 0.066\nosig
& 0.863\sig & 0.986\sig & 1.000\nosig & 1.000\nosig
& 0.742\sig & 0.921\nosig & 0.984\nosig & 0.994\nosig \\

CE+TS+TTA 
& \textbf{0.393}\nosig & \textbf{0.274}\nosig & \textbf{0.157}\nosig & \textbf{0.067}\nosig
& 0.879\sig & 0.988\sig & 1.000\nosig & 1.000\nosig
& 0.758\nosig & \textbf{0.922}\nosig & 0.984\nosig & 0.995\nosig \\
\bottomrule
\end{tabular}
\label{tab:uncertain-volume-metrics}
\end{table*}

%\begin{figure*}
%    \centering
%    \includegraphics[width=0.7\textwidth]{figs/4_results/Uncertainty_qualitative_results_shortened.pdf}
%    \caption{Comparison of predictive uncertainty maps (left column) and segmentation results (right column) across three anatomical planes (coronal, sagittal, axial) for \texttt{BASE}, \texttt{CE}, and \texttt{CE+TS}. The right column shows the predicted segmentation mask (blue) and the ground truth mask (red) overlaid on the original image.}
%    \label{fig:qualitative_results}
%\end{figure*}

\subsection{Uncertainty visualization and clinical integration}
To obtain clinically actionable uncertainty visualizations, it might be useful to \emph{threshold} the entropy map, i.e., highlighting only regions above an operating point. 
Such operating point can be selected (i) to match a \emph{revision budget} (e.g., the top-$b\%$ most uncertain voxels, consistent with the percentile-based analysis used for the AUC metrics), or (ii) as a fixed threshold optimized for other criteria (e.g., maximizing uncertainty-error alignment on a validation set). 
For enhanced visualization, we also rescale the retained uncertainty values to $[0,1]$ (voxels below threshold are set to zero/transparent), improving contrast while preserving the spatial extent implied by the selected budget.

\begin{figure*}
    \centering
    \includegraphics[width=0.85\textwidth]{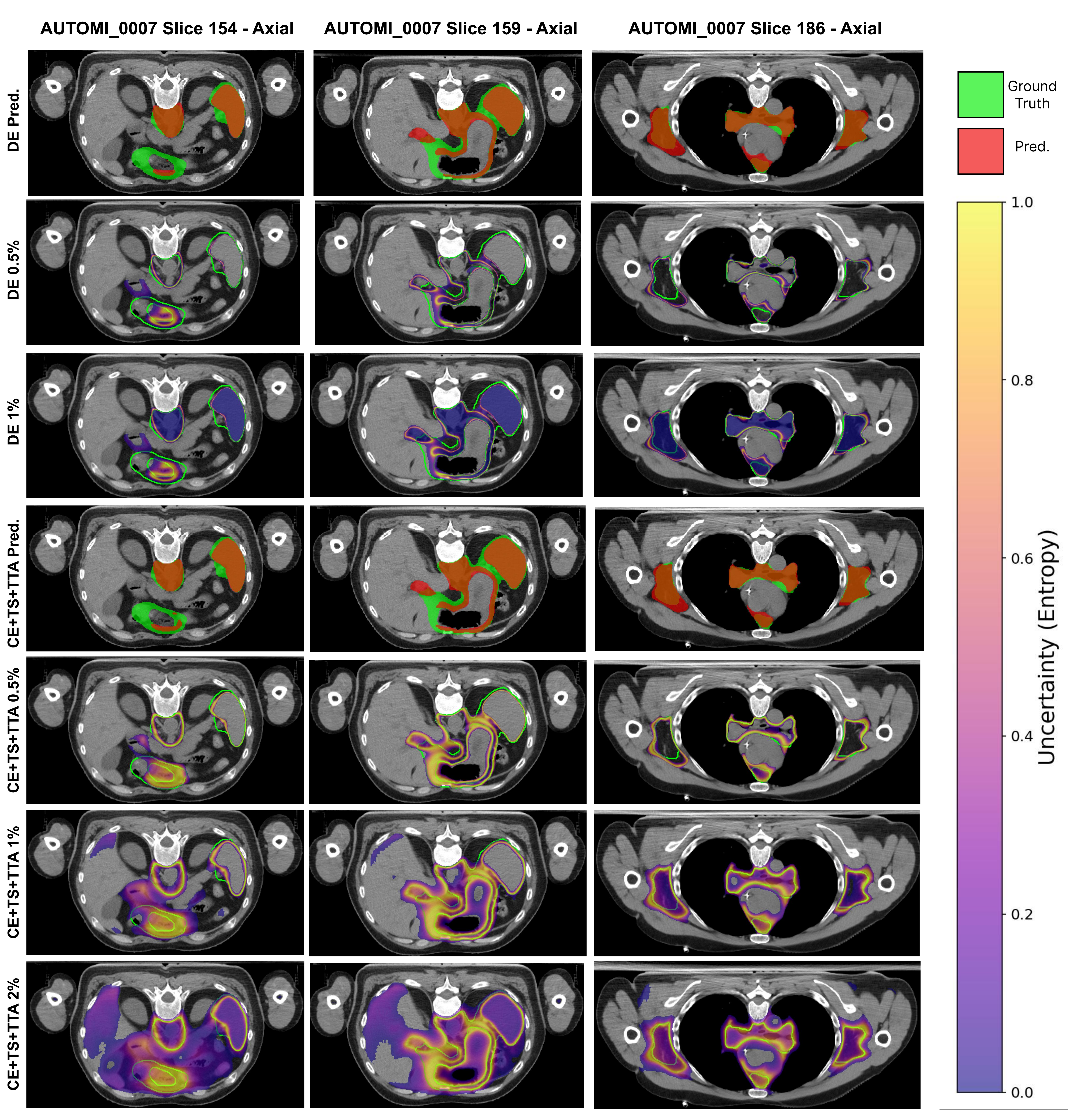}
    \caption{Budget-aware uncertainty visualization on three axial slices from a representative case (AUTOMI\_0007), comparing DE and CE+TS+TTA.
    The first and fourth rows overlay the filled ground truth (green) and prediction (red).
    Subsequent rows show entropy-based uncertainty maps thresholded by uncertain-volume budgets (top-0.5\%, 1\%, and 2\% most uncertain voxels), displayed with the ground-truth contour (green). \texttt{DE} at 2\% is omitted since not informative, i.e., $<1\%$ of voxels have non-zero uncertainty.}
    \label{fig:volume_percentage_visualization}
\end{figure*}

Figure~\ref{fig:volume_percentage_visualization} shows three axial slices and compares two uncertainty estimation strategies (DE vs.\ CE+TS+TTA) across uncertain-volume budgets of 0.5\%, 1\%, and 2\%. For each slice, the first and fourth rows overlay the \emph{filled} ground-truth mask (green) and the model prediction (red) to localize gross mismatches, while the subsequent rows show the \emph{thresholded} entropy maps at different budgets together with the ground-truth contour (green) to facilitate interpretation of uncertainty relative to anatomical boundaries.
In these examples, we can see clearly how increasing the budget from 0.5\% to 2\% progressively expands the highlighted regions, naturally supporting a \emph{budget-aware} workflow in which the clinician starts from a small, high-priority subset and incrementally inspects larger uncertain regions until the expected marginal gain no longer justifies additional review time.
Moreover, we also observe the qualitative differences between the two uncertainty estimation methods:
at 0.5\%, \texttt{DE} predominantly highlights boundary uncertainty; at 1\%, the highlighted entropy spreads broadly within the predicted positive mask, with comparatively limited focus on localized discrepancies; consequently, false positives are not clearly distinguished and missed regions (false negatives) are only weakly emphasized in these examples. We omit \texttt{DE} at 2\% because the entropy values are highly skewed toward zero; consequently, the 2\% operating point collapses to a zero threshold and provides no additional informative uncertain region beyond 1\%.
In contrast, CE+TS+TTA produces more structured and spatially coherent uncertain regions that better capture clinically relevant errors (e.g., missed lymph-node/axillary regions) already at 0.5\%, and expands more gradually with increasing budget.
Overall, this qualitative evidence complements our budget-wise quantitative findings by illustrating why combining checkpoint ensembling with calibration and augmentation can yield more \emph{actionable} uncertainty maps for segmentation quality assurance.

\section{Conclusions}
\label{sec:conclusions}
This study investigated uncertainty-driven quality QA for automatic target segmentation in radiotherapy, with the explicit goal of supporting \emph{budget-limited} manual review rather than maximizing segmentation accuracy alone. Across all evaluated configurations, segmentation performance remained stable (DSC in the range 0.836--0.843), indicating that UQ and post-hoc calibration can be introduced without materially changing the underlying hard segmentation.

TS improved probabilistic calibration at negligible cost (single-scalar fitting on validation data, no retraining). For the baseline, ROI-masked ECE decreased from $0.062 \pm 0.015$ to $0.031 \pm 0.006$ (BASE vs.\ BASE+TS) with unchanged DSC; similar trends were observed when TS was combined with ensembling.

Under realistic revision constraints, UQ improved \emph{uncertainty--error alignment}. In the budget-wise analysis over the top 0--5\% most uncertain voxels, CE achieved the strongest overlap, with CE+TS+TTA yielding the best UEO$_{\mathrm{AUC}}$ (0.173 vs.\ 0.154 for the baseline). Coverage-separation results further suggest configuration-specific trade-offs: CE-based variants preferentially highlighted false positives (best (FP--TP)$_{\mathrm{AUC}}$ with CE+TTA), while DE+TS+TTA maximized false-negative separation ((FN--TN)$_{\mathrm{AUC}}$ up to 0.900 vs.\ 0.674 for the baseline). At tight budgets, CE+TS+TTA also achieved the highest point-wise UEO at 0.5\% uncertain volume.

Overall, these results support a \emph{hierarchical, budget-driven QA} workflow in which the reviewer starts from the top 0.5--1\% most uncertain voxels and expands only if needed. In the presented examples, CE+TS+TTA produced spatially coherent uncertainty regions that better matched clinically relevant corrections than DE at the same budget. Practically, CE is attractive for deployment because it reuses checkpoints from a single training run, avoiding the cost of training multiple independent models.

Despite these findings, our study is preliminary and has limitations that should be addressed in future work. First, all experiments were conducted on a single-center in-house cohort ($n=45$), and validation on larger \emph{multicenter} datasets is needed to assess generalizability across scanners, contouring guidelines, and patient populations. Second, key hyperparameters (e.g., the temperature $T$, the number/timing of checkpoints, and the number/type of TTA augmentations) were not systematically optimized. Finally, the retrospective design limits the assessment of the real-world impact of uncertainty-guided review on clinical practice. Future work will involve a prospective, blinded evaluation in which uncertainty-guided review is compared against standard practice using (i) edit-mask--based endpoints (e.g., overlap between flagged voxels and actual edits) and (ii) clinically meaningful efficiency measures such as correction time and inter-observer variability.

\section*{Acknowledgment}
The authors disclose the use of OpenAI ChatGPT (version 5.2) to assist with English-language editing and linguistic refinement of the manuscript. The use of AI was limited to language polishing, while all technical and scientific content was produced and verified by the authors.

% This work is part of the AuToMI project funded by the Italian Ministry of Health (Rome, Italy; grant number: GR-2019-12370739).

\bibliographystyle{ieeetr} 
\bibliography{myrefs.bib}

%\appendices
%\input{appendix}

\end{document}